\documentclass{article}
\usepackage{url}
\usepackage[preprint]{neurips_2024}
\usepackage[utf8]{inputenc}
\usepackage{amsmath,amssymb, amsthm}
\usepackage{algorithm,algpseudocode}
\usepackage{wrapfig}
\usepackage{graphicx}
\usepackage{booktabs}
\usepackage{hyperref}
\title{Nonnegative Matrix Factorization through Cone Collapse}
\author{
Manh Nguyen\\
Department of Statistics\\
Wisconsin Institute of Discovery\\
University of Wisconsin-Madison\\
\texttt{mdnguyen4@wisc.edu}
\And
Daniel Pimentel-Alarc\'on\\
Department of Biostatistics and Medical Informatics\\
Wisconsin Institute of Discovery\\
University of Wisconsin-Madison\\
\texttt{pimentelalar@wisc.edu}
}
\begin{document}
\maketitle

\newtheorem{theorem}{Theorem}[section]
\newtheorem{lemma}[theorem]{Lemma}
\newtheorem{proposition}[theorem]{Proposition}
\newtheorem{corollary}[theorem]{Corollary}

\theoremstyle{definition}
\newtheorem{definition}[theorem]{Definition}

\newtheorem{assumption}[theorem]{Assumption}

\theoremstyle{remark}
\newtheorem{remark}[theorem]{Remark}

\begin{abstract}
Nonnegative matrix factorization (NMF) is a widely used tool for learning
parts-based, low-dimensional representations of nonnegative data, with
applications in vision, text, and bioinformatics
\cite{lee1999nmf,lee2000nmf}. In clustering applications, orthogonal
NMF (ONMF) variants further impose (approximate) orthogonality on the
representation matrix so that its rows behave like soft cluster
indicators \cite{ding2005equiv,ding2006orthnmf}. Existing algorithms,
however, are typically derived from optimization viewpoints and do not
explicitly exploit the conic geometry induced by NMF: data points lie in
a convex cone whose extreme rays encode fundamental directions or
"topics". In this work we revisit NMF from this geometric perspective
and propose \emph{Cone Collapse}, an algorithm that starts from the full
nonnegative orthant and iteratively shrinks it toward the minimal cone
generated by the data. We prove that, under mild assumptions on the data,
Cone Collapse terminates in finitely many steps and recovers the minimal
generating cone of $\mathbf{X}^\top$. Building on this basis, we then
derive a cone-aware orthogonal NMF model (CC--NMF) by applying
uni-orthogonal NMF to the recovered extreme rays
\cite{ding2006orthnmf,yoo2008onmf}. Across 16 benchmark gene-expression,
text, and image datasets, CC--NMF consistently matches or outperforms
strong NMF baselines---including multiplicative updates, ANLS, projective
NMF, ONMF, and sparse NMF---in terms of clustering purity. These results
demonstrate that explicitly recovering the data cone can yield both
theoretically grounded and empirically strong NMF-based clustering
methods. The implementation for our method is provided in \href{https://github.com/manhbeo/cone-collapse}{github.com/manhbeo/cone-collapse}.
\end{abstract}

\section{Introduction}

Low-rank matrix factorization methods are central tools for discovering
low-dimensional structure in high-dimensional data. Classical techniques
such as principal component analysis (PCA) and singular value
decomposition (SVD) provide optimal rank-$r$ approximations in the
least-squares sense, but their components typically contain both
positive and negative entries, which complicates interpretation when the
data are inherently nonnegative (e.g., pixel intensities, word counts,
gene-expression levels). Nonnegative matrix factorization (NMF)
\cite{lee1999nmf,lee2000nmf} addresses this limitation by constraining
both factors to be nonnegative. The resulting additive, parts-based
decompositions have been successfully used for document clustering
\cite{xu2003nmfclust}, topic modeling, and molecular pattern discovery,
and have been linked to probabilistic latent semantic indexing and
related latent-variable models \cite{ding2005equiv}.

Beyond representation learning, NMF has a long-standing connection to
clustering. By constraining one factor to be close to an indicator
matrix, NMF can be shown to approximate $k$-means and spectral
clustering objectives \cite{ding2005equiv}. Orthogonal NMF (ONMF)
formulations make this connection explicit by enforcing an orthogonality
constraint on either the basis or the coefficient matrix, e.g.,
$\mathbf{H}\mathbf{H}^\top = \mathbf{I}$ in a factorization
$\mathbf{X}\approx \mathbf{W}\mathbf{H}$.
In such models, each row of $\mathbf{H}$ behaves like a soft indicator
vector for a cluster, and assigning each data point to the row with
maximal activation yields a clustering
\cite{ding2006orthnmf,choi2008onmf,yoo2008onmf,pompili2013onpmf}.
These ONMF variants have been particularly successful on document and
image clustering benchmarks.

Most NMF and ONMF algorithms are derived from an optimization viewpoint,
using multiplicative updates \cite{lee2000nmf}, projected gradient
methods, or alternating nonnegative least squares (ANLS) with advanced
solvers such as block principal pivoting \cite{kim2008faster}. While
effective in practice, these approaches rarely exploit the explicit
\emph{conic} geometry underlying NMF: the columns of a data matrix
$\mathbf{X}$ lie in the convex cone $\operatorname{cone}(\mathbf{W})$
generated by the basis vectors, and the extreme rays of this cone play a
role analogous to “topics'' or “anchors''. In parallel, a line of work
on separable NMF and topic modeling has developed algorithms that
directly recover extreme rays (or “anchor words'') under structural
assumptions, with provable guarantees
\cite{arora2013topic,ding2005equiv}. However, these methods typically
operate in simplex or probability-simplex settings and are not designed
to integrate with ONMF-style clustering objectives.

\paragraph{Our approach.}
In this paper we propose \emph{Cone Collapse}, a new algorithm that
explicitly recovers the minimal generating cone of the data and then
uses it as the basis for an ONMF-style clustering model. Given a
nonnegative data matrix $\mathbf{X}\in\mathbb{R}^{m\times n}_+$, we
view its transpose $\mathbf{X}^\top$ as a set of points in
$\mathbb{R}^n_+$ and seek a matrix
$\mathbf{U}^\star=[\mathbf{u}_1,\dots,\mathbf{u}_c]$ whose columns
correspond to extreme rays of the data cone
$\operatorname{cone}(\mathbf{X}^\top)$. Cone Collapse starts from the
full nonnegative orthant (via the identity matrix) and iteratively
\emph{shrinks} this cone by tilting free rays toward the mean direction
of the data while ensuring that all points remain inside the cone. When
a data point falls outside, it is added as a new ray; when a ray becomes
representable as a nonnegative combination of others, it is pruned.
These steps are implemented via NNLS subproblems solved efficiently by
block principal pivoting \cite{kim2008faster}. Intuitively, the
algorithm contracts an initial, overly large cone until only the
essential extreme rays remain.

\begin{figure}
    \centering
    \includegraphics[width=1\linewidth]{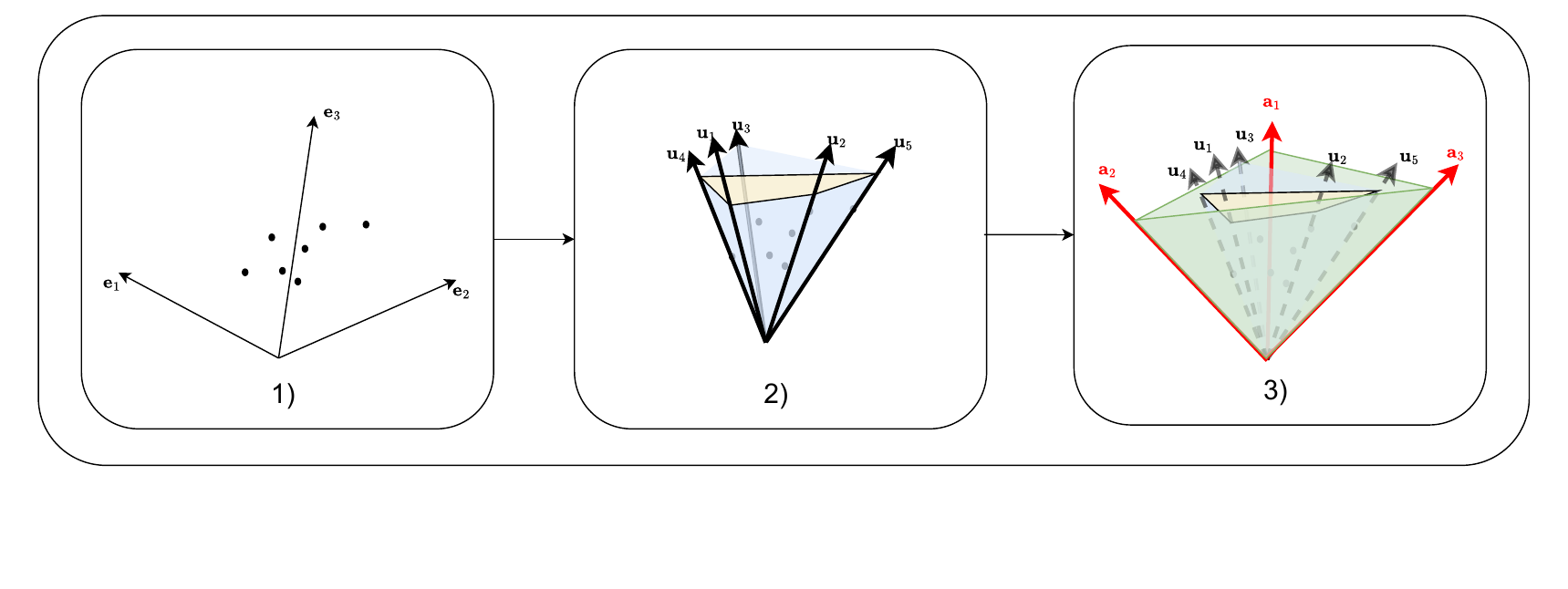}
    \caption{An illustration of our method: (1) initialize with basis vectors $e_1, e_2, e_3$; (2) recover the data cone $\mathbf{U}^\star$ that contains all columns of $\mathbf{X}^\top$ via Cone Collapse; and (3) fit an orthogonal cone $\mathbf{A}$ to $\mathbf{U}^\star$.}
    \label{fig:process}
\end{figure}

Once $\mathbf{U}^\star$ has been recovered, we fit another
orthogonal cone $\operatorname{cone}(\mathbf{A})$ of $r$ rays to
$\operatorname{cone}(\mathbf{U}^\star)$ by solving a uni-orthogonal
NMF problem $\mathbf{U}^\star \approx \mathbf{A}\mathbf{S}$ with
$\mathbf{A}^\top\mathbf{A}=\mathbf{I}$ using multiplicative ONMF
updates \cite{ding2006orthnmf,yoo2008onmf}. This yields an orthogonal
factorization $\mathbf{X} \approx \mathbf{W}\mathbf{H}$, where
$\mathbf{H}=\mathbf{A}^\top$ has orthonormal rows and can be used
directly for clustering. We refer to the resulting model as
\emph{CC--NMF} (Cone Collapse NMF). Figure~\ref{fig:process} provides a
schematic illustration of this two-stage pipeline.

\paragraph{Contributions.}
Our main contributions are:
\begin{itemize}
\item We introduce \textbf{Cone Collapse}, a new algorithm for recovering
      a minimal generating cone of a nonnegative data matrix. The method
      combines mean-tilting, outside-point detection, and redundancy
      pruning, and is built on efficient NNLS routines
      \cite{kim2008faster}.
\item We provide a \textbf{theoretical justification} for Cone Collapse:
      under mild assumptions (clean data and nondegenerate extreme
      rays), we prove that the algorithm terminates after finitely many
      iterations and returns a basis whose cone coincides with the data
      cone, i.e., $\operatorname{cone}(\mathbf{U}^{(T)})
      = \operatorname{cone}(\mathbf{X}^\top)$.
\item We show how to integrate Cone Collapse with \textbf{orthogonal
      NMF}, yielding CC--NMF, a cone-aware ONMF model whose latent
      factors are explicitly tied to extreme rays of the data cone, in
      contrast to existing ONMF formulations
      \cite{ding2006orthnmf,choi2008onmf,yoo2008onmf,pompili2013onpmf}.
\item We conduct a \textbf{comprehensive empirical evaluation} on 16
      benchmark datasets spanning gene expression, text, and images, and
      demonstrate that CC--NMF consistently matches or outperforms
      strong NMF baselines---including MU, ANLS, PNMF, ONMF, and sparse
      NMF---in clustering purity.
\end{itemize}

\paragraph{Organization.}
Section~\ref{sec:related-work} reviews related work on NMF and ONMF.
Section~\ref{sec:method} introduces the Cone Collapse algorithm and its
geometric interpretation. Section~\ref{sec:theory} establishes the
finite-termination and exact-recovery guarantees. Section~\ref{sec:experiments}
describes our experimental setup and clustering results, and
Section~\ref{sec:conclusion} concludes with a discussion of limitations
and future directions.

\section{Related work}\label{sec:related-work}
\paragraph{Matrix factorization and parts-based representations.}
Matrix factorization methods such as singular value decomposition (SVD) and principal
component analysis (PCA) have long been used to obtain low-dimensional representations
of high-dimensional data. However, these factorizations typically produce components with
both positive and negative entries, which hinders interpretability when the data themselves
are nonnegative (e.g., images, word counts, or term–document matrices). Nonnegative
matrix factorization (NMF) addresses this issue by constraining both factors to be
nonnegative, leading to parts-based or additive representations that have proven useful in
computer vision, text analysis, and bioinformatics \cite{lee1999nmf,lee2000nmf}. Subsequent
work has revealed close connections between NMF and clustering objectives, including
equivalences to $k$-means and spectral clustering under appropriate constraints
\cite{ding2005equiv}.

\paragraph{Nonnegative matrix factorization.}
Nonnegative matrix factorization (NMF) seeks to approximate a nonnegative data matrix 
$\mathbf{X} = [\mathbf{x}_1,\dots,\mathbf{x}_n]\in\mathbb{R}^{m\times n}_+$ by a low–rank product
\[
\mathbf{X} \;\approx\; \mathbf{W}\mathbf{H},\qquad
\mathbf{W} \in \mathbb{R}^{m\times r}_+,\ \mathbf{H} \in \mathbb{R}^{r\times n}_+.
\]
We assume $\mathbf{X}$ is clean, i.e.\ there is no full zero column or row in $\mathbf{X}$.
Here, $n$ is the number of examples and $m$ is the number of features. Each column
$\mathbf{x}_i$ is then represented as a nonnegative combination of the basis vectors
$\mathbf{w}_1,\dots,\mathbf{w}_r$ (the columns of $\mathbf{W}$), i.e.
\[
\mathbf{x}_i \;\approx\; \mathbf{W}\mathbf{h}_i \;=\; \sum_{k=1}^r h_{ki}\,\mathbf{w}_k,\qquad h_{ki}\ge 0.
\]
Geometrically, this means that all data points $\mathbf{x}_i$ lie (approximately) inside the convex cone
generated by the columns of $\mathbf{W}$,
\[
\operatorname{cone}(\mathbf{W})
\;:=\;
\Big\{\sum_{k=1}^r \alpha_k \mathbf{w}_k : \alpha_k \ge 0\Big\}.
\]
Thus, NMF can be interpreted as the problem of finding a low–dimensional cone that
captures the data cloud $\mathbf{X}$ \cite{lee2000nmf}.

\paragraph{Orthogonal nonnegative matrix factorization for clustering.}
Orthogonal nonnegative matrix factorization (ONMF) augments NMF with an orthogonality
constraint on one of the factors, typically
\[
\mathbf{X} \;\approx\; \mathbf{W}\mathbf{H},\qquad
\mathbf{W} \in \mathbb{R}^{m\times r}_+,\ \mathbf{H} \in \mathbb{R}^{r\times n}_+,
\quad \mathbf{H}\mathbf{H}^\top = \mathbf{I}_r,
\]
or analogously with orthogonality imposed on the columns of $\mathbf{W}$. The
nonnegativity of $\mathbf{H}$ encourages each data point $\mathbf{x}_i$ to be represented
by a small number of latent components, while the orthogonality constraint forces the rows
of $\mathbf{H}$ to behave like (soft) cluster-indicator vectors. A common clustering
interpretation is to assign each data point $\mathbf{x}_i$ to the cluster
\[
\operatorname*{arg\,max}_{k\in\{1,\dots,r\}} h_{ki},
\]
so that ONMF plays the role of a relaxed combinatorial clustering formulation. Under
suitable conditions, these orthogonality constraints make NMF and its symmetric variants
equivalent to $k$-means or spectral clustering objectives
\cite{ding2005equiv,ding2006orthnmf}, thereby providing a principled link between
matrix factorization and graph-based clustering. While our approach is broadly applicable to NMF problems, in this work we focus on combining Cone Collapse with orthogonal NMF and evaluating it on clustering tasks.

\section{Method}\label{sec:method}
\subsection{Cone Collapse Algorithm}
\begin{wrapfigure}{r}{0.5\textwidth}
    \includegraphics[width=1\linewidth]{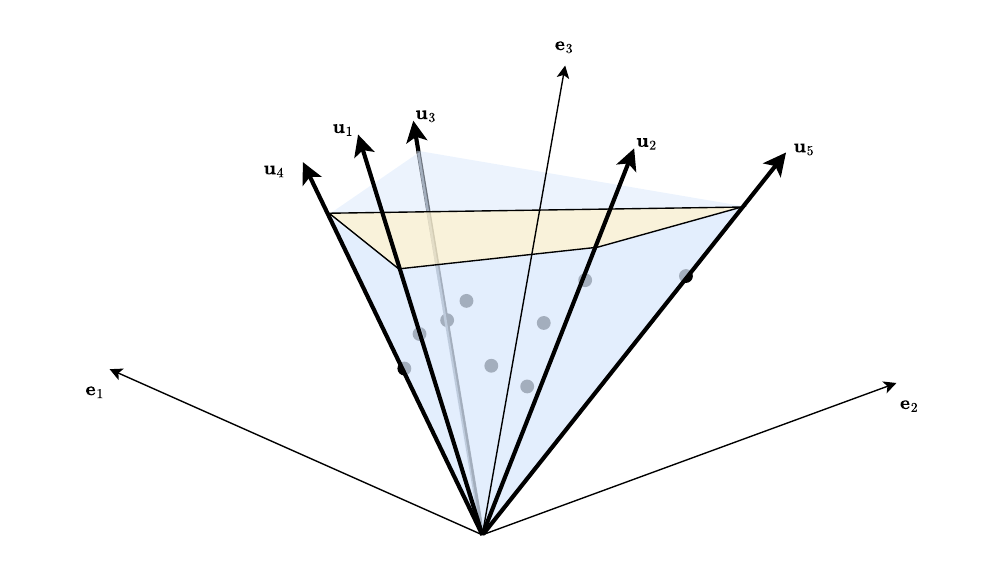}
    \caption{A 3D illustration for $\mathbf{U}^\star$ with $c = 6$. Each column of $\mathbf{U}^\star$ is represented as a ray, with the cone $\operatorname{cone}(\mathbf{U}^\star)$ contains all datapoints of $\mathbf{X}^\top$ (which are rows of $\mathbf{X}$, represented as points).}
    \label{fig:U}
\end{wrapfigure} Our goal in the first step is to recover a convex cone 
$\mathbf{U}^\star=[\mathbf{u}_1,\dots,\mathbf{u}_c]\in\mathbb{R}^{m\times c}$, whose generating rays capture the geometry of the data matrix 
$\mathbf{X}^\top$. Geometrically, the columns of $\mathbf{U}^\star$ correspond to (a
subset or superset of) the \emph{extreme rays} of the data cone
\[
\operatorname{cone}(\mathbf{X}^\top)
\;:=\;
\{\mathbf{X}^\top\alpha : \alpha \ge 0\}.
\]

A ray $\mathbb{R}_+\mathbf{u} \subset \operatorname{cone}(\mathbf{X}^\top)$ is called
\emph{extreme} if it cannot be written as a nontrivial conic combination of other
rays in the cone: whenever $\mathbf{u} = \mathbf{v} + \mathbf{w}$ with 
$\mathbf{v},\mathbf{w}\in\operatorname{cone}(\mathbf{X}^\top)$, we must have 
$\mathbf{v} = a\mathbf{u}$ and $\mathbf{w} = b\mathbf{u}$ for some $a,b \ge 0$.
In other words, extreme rays play the role of "corners" of the cone. An illustration for $\mathbf{U}^\star$ is provided in Figure \ref{fig:U}. If $c = 1$, the cone reduces to a single ray and the problem is trivial: any nonzero rows of $\mathbf{X}$, after normalization, provides the unique direction. Therefore, we focus on the nontrivial regime $c \ge 2$. 

\paragraph{Intuition.}
The guiding idea behind our method is simple: we begin with a cone so large that it already 
contains all data points, and then we continuously \emph{shrink} this cone toward a compact shape that reveals the true extreme rays of the data. Specifically, we initialize with $U^{(0)} = I_n$, which guarantees that every data point $x_i$ lies inside 
$\operatorname{cone}(U^{(0)})$. From this starting point, we iteratively ``tilt'' each ray 
$\mathbf{u}_k^{(t)}$ toward the mean direction $\mu$ of the dataset. During this contraction process some data points may fall outside the current cone, which is then added to the cone. Finally, we prune any rays that have become redundant—those that can be expressed as a 
nonnegative combination of others—ensuring that the cone remains minimal.

For nonzero vectors $\mathbf{a}$ and $\mathbf{b}$, denote $\widehat{\mathbf{a}} := \mathbf{a}/\|\mathbf{a}\|$ and $\cos(\mathbf{a},\mathbf{b}) := \langle \widehat{\mathbf{a}},\widehat{\mathbf{b}}\rangle$; the Cone Collapse algorithm is describe in Algorithm \ref{alg:cone_collapse}.

\begin{algorithm}[t]
\caption{Cone–Collapse Algorithm}
\label{alg:cone_collapse}
\begin{algorithmic}[1]
\Require Data $\mathbf{X}^\top=[\mathbf{x}_1,\dots,\mathbf{x}_m]\in\mathbb{R}^{n\times m}_+$, learning rate $\eta\in(0,1)$, tolerance $\epsilon$.
\Ensure Final basis $\mathbf{U}^{(T)}$
\State \textbf{init }$\boldsymbol{\mu} \gets \dfrac{1}{m}\sum_{i=1}^m \mathbf{x}_i$, \quad $\mathbf{U}^{(0)} \gets \mathbf{I}_m$, \quad $t \gets 0$
\Repeat \Comment{Iteration $t$}
    \For{each column $\mathbf{u}_k^{(t)}$ of $\mathbf{U}^{(t)}$}
     \Comment{\emph{Mean tilt}}
        \[
        \widetilde{\mathbf{u}}_k \leftarrow
        \begin{cases}
        \mathbf{u}_k^{(t)}, & \text{If } \exists\, i\in[n]: \cos(\mathbf{u}_k^{(t)},\mathbf{x}_i)=1,\\
        \dfrac{(1-\eta)\,\mathbf{u}_k^{(t)}+\eta\,\widehat{\boldsymbol{\mu}} }
              {\|(1-\eta)\,\mathbf{u}_k^{(t)}+\eta\,\widehat{\boldsymbol{\mu}} \|_2}, & \text{otherwise.}
        \end{cases}
        \]
    \EndFor
    \State $\widetilde{\mathbf{U}}^{(t)} \gets [\widetilde{\mathbf{u}}_1,\dots,\widetilde{\mathbf{u}}_{c_t}]$ \Comment{$c_t$: number of columns of $\mathbf{U}^{(t)}$}
    \State $\mathbf{H}^\star \gets \arg\min_{\mathbf{H}\ge 0}\ \big\|\mathbf{X}^\top-\widetilde{\mathbf{U}}^{(t)} \mathbf{H}\big\|_F^2$
    \State $\mathbf{R} := [\mathbf{r}_1,\dots,\mathbf{r}_n]  \gets \mathbf{X}^\top-\widetilde{\mathbf{U}}^{(t)} \mathbf{H}^\star$
    \For{$i = 1,\dots,n$}
        \If{$\|\mathbf{r}_i\|_2 > \epsilon\,\|\mathbf{x}_i\|_2$}
            \State $\widetilde{\mathbf{U}}^{(t)} \gets \big[\,\widetilde{\mathbf{U}}^{(t)}\ \ \widehat{\mathbf{x}}_i\,\big]$
            \Comment{\emph{Add outside point}}
            \For{each column index $k$ of $\widetilde{\mathbf{U}}^{(t)}$}
                \State $\widetilde{\mathbf{U}}^{(t)}_{-k} \gets (\widetilde{\mathbf{U}}^{(t)}$ with column $k$ removed)
                \State $\mathbf{w}_k^\star \gets \arg\min_{\mathbf{w}\ge 0}\ \big\|\widetilde{\mathbf{u}}_k-\widetilde{\mathbf{U}}^{(t)}_{-k} \mathbf{w}\big\|_2^2$, \quad $\rho_k \gets \big\|\widetilde{\mathbf{u}}_k-\widetilde{\mathbf{U}}^{(t)}_{-k} \mathbf{w}_k^\star\big\|_2$
                \If{$\rho_k \le \epsilon\,\|\widetilde{\mathbf{u}}_k\|_2$}
                    \State $\widetilde{\mathbf{U}}^{(t)} \gets \widetilde{\mathbf{U}}^{(t)}_{-k}$
                    \Comment{\emph{Remove redundant rays}}
                \EndIf
            \EndFor
        \EndIf
    \EndFor
    \State $\mathbf{U}^{(t+1)} \gets \widetilde{\mathbf{U}}^{(t)}$
    \State $t \gets t+1$
\Until{$\forall\, k\ \exists\, i\in[n] \text{ such that } \cos(\widetilde{\mathbf{u}}_k,\mathbf{x}_i)=1$}
\State \Return $\mathbf{U}^{(T)} \gets \mathbf{U}^{(t)}$
\end{algorithmic}
\end{algorithm}

\textbf{Discussion:} 
\begin{itemize}
    \item The \emph{Mean tilt} step leaves the columns of $\widetilde{\mathbf{U}}^{(t)}$ that are co-linear with a point in $\mathbf{X}^\top$ unchanged and tilts the columns that are not co-linear with any point in $\mathbf{X}$ by
    \[
        \mathbf{u}^{(t)}_k \;\mapsto\; \widetilde{\mathbf{u}}^{(t)}_k
        \propto (1-\eta)\,\mathbf{u}^{(t)}_k + \eta\,\widehat{\boldsymbol{\mu}}.
    \]
    All columns always remain in $\mathbb{S}^{n-1}_+ = \{\mathbf{v}\in\mathbb{R}^n_+:\|\mathbf{v}\|_2=1\}$.

    \item After each iteration $t$ of the algorithm, we have $\mathbf{X}^\top\subseteq \operatorname{cone}\big(\mathbf{U}^{(t+1)}\big)$. Indeed, at the \emph{Add outside point} step every $\mathbf{x}_i$ with residual $\|\mathbf{r}_i\|_2> \epsilon\|\mathbf{x}_i\|_2$ is appended as $\widehat{\mathbf{x}}_i$, so
    $\mathbf{X}^\top\subseteq \operatorname{cone}(\widetilde{\mathbf{U}}^{(t)})$ holds immediately after that step. The \emph{Remove redundant rays} step only removes columns
    $\widetilde{\mathbf{u}}_k$ that satisfy $\widetilde{\mathbf{u}}_k\in \operatorname{cone}(\widetilde{\mathbf{U}}^{(t)}_{-k})$ (up to $\epsilon$), so deleting
    them does not change the cone.

    \item $\arg\min_{\mathbf{w}\ge 0}\ \big\|\widetilde{\mathbf{u}}_k-\widetilde{\mathbf{U}}^{(t)}_{-k} \mathbf{w}\big\|_2^2$ and $\mathbf{H}^\star \gets \arg\min_{\mathbf{H}\ge 0}\ \big\|\mathbf{X}^\top-\widetilde{\mathbf{U}}^{(t)} \mathbf{H}\big\|_F^2$ are solved by Algorithm~\ref{alg:bpp-single} and Algorithm~\ref{alg:bpp-multi} provided in Appendix D. One may consider adding all the outside points before removing redundant rays; however, in practice with large $m$ and $n$, this can increase the solving time and lead to unnecessary NNLS calls.

    \item The learning rate $\eta$ does not affect the results of the algorithm, but it affect the time required to converge. A smaller $\eta$ typically requires more iterations but leads to a smaller number of points falling outside $\operatorname{cone}(\mathbf{U}^{(t)})$ and vise versa. 
\end{itemize}

\subsection{Orthogonal NMF using Cone Collapse algorithm}
Recall that our goal is to obtain an orthogonal NMF of the form
\[
\mathbf{X} \;\approx\; \mathbf{W}\mathbf{H},\qquad
\mathbf{W} \in \mathbb{R}^{m\times r}_+,\ \mathbf{H} \in \mathbb{R}^{r\times n}_+,\ \mathbf{H}\mathbf{H}^\top = \mathbf{I}_r,
\]
so that the rows of $\mathbf{H}$ act as (approximately) orthogonal cluster indicators for the columns of $\mathbf{X}$, in the spirit of orthogonal NMF and its connection to $k$-means clustering \cite{ding2006orthnmf}.

After obtaining the extreme–ray basis $\mathbf{U}^\star \in \mathbb{R}_+^{n\times c}$ from Algorithm~\ref{alg:cone_collapse}, we first solve a nonnegative least–squares problem
\[
\mathbf{V}^\star
\;=\;
\arg\min_{\mathbf{V}\ge 0}\ \big\|\mathbf{X}^\top- \mathbf{U}^\star \mathbf{V}\big\|_F^2,
\]
so that $\mathbf{X}^\top \approx \mathbf{U}^\star \mathbf{V}^\star$ and hence $\mathbf{X} \approx (\mathbf{V}^\star)^\top (\mathbf{U}^\star)^\top$.

We then compress the cone basis $\mathbf{U}^\star$ by solving a uni-orthogonal ONMF problem
\[
\min_{\mathbf{A}\ge 0,\ \mathbf{S}\ge 0}
\ \big\|\mathbf{U}^\star - \mathbf{A}\mathbf{S}\big\|_F^2
\quad\text{s.t.}\quad
\mathbf{A}^\top \mathbf{A} = \mathbf{I}_r,
\]
where $\mathbf{A}\in\mathbb{R}_+^{n\times r}$ has orthonormal columns and $\mathbf{S}\in\mathbb{R}_+^{r\times c}$ contains nonnegative loadings. Note that all columns of $\mathbf{U}^\star$ are in $\operatorname{cone}(\mathbf{A})$, so geometrically we are fitting another cone with a smaller number of extreme rays $r \le c$ to the existing $\operatorname{cone}(\mathbf{U}^\star)$.

Following the uni-orthogonal NMF updates of Ding et al.\ \cite{ding2006orthnmf}, we adopt the multiplicative rules
\begin{align}
\mathbf{S}
&\leftarrow
\mathbf{S}\ \odot\
\frac{\mathbf{A}^\top \mathbf{U}^\star}{(\mathbf{A}^\top \mathbf{A})\,\mathbf{S}},
\label{eq:onmf-S-update}
\\[3pt]
\mathbf{A}
&\leftarrow
\mathbf{A}\ \odot\
\frac{\mathbf{U}^\star \mathbf{S}^\top}{\mathbf{A}\mathbf{A}^\top \mathbf{U}^\star \mathbf{S}^\top},
\label{eq:onmf-A-update}
\end{align}
where all products/divisions are taken elementwise. In practice we periodically re-normalize the columns of $\mathbf{A}$ to keep $\mathbf{A}^\top \mathbf{A}\approx \mathbf{I}_r$, as is standard in ONMF algorithms based on multiplicative updates on the Stiefel manifold \cite{yoo2008onmf,choi2008onmf}.

Combining the two stages, we obtain the overall approximation
\[
\mathbf{X}
\;\approx\;
(\mathbf{V}^\star)^\top \mathbf{S}^\top \mathbf{A}^\top,
\]
so that a valid ONMF of $\mathbf{X}$ is given by
\[
\mathbf{W} := (\mathbf{V}^\star)^\top \mathbf{S}^\top \in \mathbb{R}_+^{m\times r},
\qquad
\mathbf{H} := \mathbf{A}^\top \in \mathbb{R}_+^{r\times n},
\]
and the orthogonality constraint holds as
\[
\mathbf{H}\mathbf{H}^\top
= \mathbf{A}^\top \mathbf{A}
= \mathbf{I}_r.
\]
Thus, Cone Collapse provides a geometrically motivated extreme-ray basis $\mathbf{U}^\star$, while the ONMF step refines it into an orthogonal low-rank factorization of $\mathbf{X}$ via the multiplicative updates \eqref{eq:onmf-S-update}–\eqref{eq:onmf-A-update}.

\section{Theoretical justification}\label{sec:theory}
In this part, we introduce a theorem (and prove it) to demonstrate why the Cone Collapse algorithm will recover the minimal generating cone of $\mathbf{X}^\top$ and terminate in finitely many iterations. Formally, let $\mathbf{U}^{(t)}=\begin{bmatrix}\mathbf{u}^{(t)}_1~\dots~\mathbf{u}^{(t)}_{c_t}\end{bmatrix}$ and define the "frozen" and "free" sets:
\[
\mathcal{F}^{(t)}:=\big\{k:\exists\,i,\ \cos(\mathbf{u}_k^{(t)},\mathbf{x}_i)=1\big\},\qquad
\mathcal{B}^{(t)}:=\{1,\dots,c_t\}\setminus\mathcal{F}^{(t)},
\]
that is, $\mathcal{F}^{(t)}$ is the set of ray indices in $\mathbf{U}^{(t)}$ that are co-linear with some data points in $\mathbf{X}^\top$, and $\mathcal{B}^{(t)}$ is the set of ray indices that are \emph{not} co-linear with any data point in $\mathbf{X}^\top$. 

We first introduce several Lemmas to help proving the theorem:

\begin{lemma}[mean direction is not an extreme ray]\label{lem:mean-not-extreme}
    Let $\mathbf{U}^\star=[\mathbf{u}_1,\dots,\mathbf{u}_c]\in\mathbb{R}^{n\times c}$ be the convex cone built from the extreme rays of $\mathbf{X}^\top$, and \(\boldsymbol{\mu} = \frac{1}{m}\sum_{i=1}^m \mathbf{x}_i\) be the mean of the data in $\mathbf{X}^\top$. If $c\ge 2$, $\forall k\in\{1,\ \dots,\ c\}$ there does not exist $\tau$ such that $\boldsymbol{\mu} = \tau\ \mathbf{u}_k$ .
    
    In other words, if there are more than 1 extreme ray, then no extreme ray is colinear with the mean.
\end{lemma}

\medskip
\begin{lemma}[contraction of free columns towards $\widehat{\boldsymbol{\mu}}$]\label{lem:tilt-contraction}
For any $\mathbf{u}^{(t)}_k\in\mathcal{B}^{(t)}$ and $\eta\in(0,1)$, if $\mathbf{u}^{(t)}_k \ne \widehat{\boldsymbol{\mu}}$
\[
\cos(\widetilde{\mathbf{u}}^{(t)}_k ,\widehat{\boldsymbol{\mu}}\big) > \cos (\mathbf{u}^{(t)}_k,\widehat{\boldsymbol{\mu}}),
\]
that is, if $\mathbf{u}^{(t)}_k \ne \widehat{\boldsymbol{\mu}}$, the cosine of the angle between $\mathbf{u}^{(t)}_k $ and the mean $\widehat{\boldsymbol{\mu}}$ strictly increases after the Mean tilt step.
\end{lemma}

\medskip
\begin{lemma}[spherical cap stability under conic combinations]\label{lem:cap-stability}
    For $\alpha\in(0,1)$, define the cap
    \begin{align*}
        \mathcal{C}_\alpha:=\{\mathbf{v}\in\mathbb{S}^{n-1}_+: \cos(\mathbf{v},\widehat{\boldsymbol{\mu}})\ge \alpha\}.
    \end{align*} 
    For any $\mathbf{b}_\ell\in\mathcal{C}_\alpha$ and $\mathbf{w} \ne 0$, if $\mathbf{w}=\sum_\ell \lambda_\ell \mathbf{b}_\ell$ with $\lambda_\ell\ge 0$ then $\widehat{\mathbf{w}}\in\mathcal{C}_\alpha$.
\end{lemma}

The proofs for the Lemmas are provided in the Appendix. Taken together, these lemmas formalize the geometric intuition behind Cone Collapse. 
Lemma~\ref{lem:mean-not-extreme} guarantees that the mean direction $\widehat{\boldsymbol{\mu}}$ is 
never itself an extreme ray, so tilting rays toward $\widehat{\boldsymbol{\mu}}$ does not accidentally 
“snap” an extreme ray into the mean. Lemma~\ref{lem:tilt-contraction} shows that every free ray 
$\mathbf{u}_k^{(t)}\in\mathcal{B}^{(t)}$ is progressively contracted toward $\widehat{\boldsymbol{\mu}}$, 
increasing its cosine with the mean at each iteration. Lemma~\ref{lem:cap-stability} then implies that, 
after sufficiently many iterations, all free rays (and any conic combination thereof) lie inside a narrow 
spherical cap around $\widehat{\boldsymbol{\mu}}$, whereas each true extreme ray can be placed outside 
this cap by an appropriate choice of its aperture. As a consequence, the algorithm is forced to add any 
missing extreme rays as new columns whenever they are detected as “outside” points, and later removes 
all non-extreme columns since they remain representable as conic combinations of the extremes. We are 
thus led to the following finite-termination and exact-recovery guarantee for Cone Collapse.

\medskip
\begin{theorem}
Algorithm \ref{alg:cone_collapse} halts after finitely many iteration $T$, with $\mathbf{U}^{(T)}$ consists of exactly $c$ columns $\{\widehat{\mathbf{u}}_1,\dots,\widehat{\mathbf{u}}_c\}$ (in some order), and
\[
 \operatorname{cone}(\mathbf{U}^{(T)})=\operatorname{cone}(\mathbf{X})=\operatorname{cone}(\mathbf{U}^\star).
\]
\end{theorem}

\emph{Proof.} To prove the theorem, we show that: (i) All extreme rays are added to $\mathbf{U}^{(t)}$ and (ii) All other columns are removed from $\mathbf{U}^{(t)}$ after finitely many iterations.

\medskip
\begin{wrapfigure}{r}{0.4\textwidth}
    \includegraphics[width=1\linewidth]{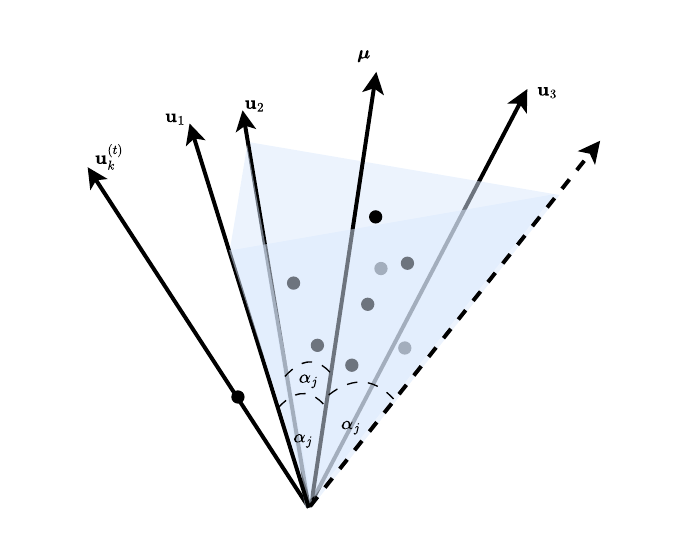}
    \caption{An illustration for $\mathcal{C}_{\alpha_i}$ contains $\operatorname{cone}\big(\{\mathbf{u}_k^{(t)}:k\in\mathcal{B}^{(t)}\}\big)$, with $\mathcal{C}_{\alpha_i}$ shaded. Here, we assume that $\mathcal{B}^{(t)} = \{1,2,3\}$. }
    \label{fig:extreme}
\end{wrapfigure}
\noindent\textbf{All extreme rays appear after finitely many iterations:} For any extreme rays $u_j$ ($j\in\{1,\ \dots,\ c\}$), suppose $\widehat{\mathbf{u}}_j$ is not added to $\mathbf{U}^{(t)}$ at the start of iteration $t$. By Lemma \ref{lem:mean-not-extreme}, we can choose $\alpha_j\in\big(\cos( \widehat{\mathbf{u}}_j,\widehat{\boldsymbol{\mu}}),1\big)$. By Lemma \ref{lem:tilt-contraction}, there exists $T_j$ such that for all $t\ge T_j$, If $\mathbf{u}^{(t)}_k\in\mathcal{B}^{(t)}$ then $ \cos (\mathbf{u}^{(t)}_k,\widehat{\boldsymbol{\mu}}) \ge \alpha_j$, which means that the $\operatorname{cone}\big(\{\mathbf{u}_k^{(t)}:k\in\mathcal{B}^{(t)}\}\big)$ is contained in $\mathcal{C}_{\alpha_j}$ by Lemma \ref{lem:cap-stability} (see Figure \ref{fig:extreme} for an illustration of $\mathcal{C}_{\alpha_i}$ contains $\operatorname{cone}\big(\{\mathbf{u}_k^{(t)}:k\in\mathcal{B}^{(t)}\}\big)$). Since $\alpha_j > \cos( \widehat{\mathbf{u}}_j,\widehat{\boldsymbol{\mu}})$, $\widehat{\mathbf{u}}_j\not\in\mathcal{C}_{\alpha_j}$, and hence $\widehat{\mathbf{u}}_j\not\in\operatorname{cone}\big(\{\mathbf{u}_k^{(t)}:k\in\mathcal{B}^{(t)}\}\big)$. 

For contradiction, we now assume that $\widehat{\mathbf{u}}_j\in\operatorname{cone}\big(\widetilde{\mathbf{U}}_k^{(t)}\big)$, which means there exists coefficients vector $\boldsymbol{\alpha} \ge 0$ such that 
\begin{align*}
     \widehat{\mathbf{u}}_j = \sum_{i\in\mathcal{B}^{(t)}} \alpha_i\mathbf{u}^{(t)}_i + \sum_{h\in\mathcal{F}^{(t)}} \alpha_h\mathbf{u}^{(t)}_h = \sum_k \alpha_k\mathbf{u}^{(t)}_k,
\end{align*}
However, by definition of extreme rays, we have $\widehat{\mathbf{u}}_j\notin\operatorname{cone}\big(\{\mathbf{u}_k^{(t)}:k\in\mathcal{F}^{(t)}\}\big)$. Hence, $ \sum_{i\in\mathcal{B}^{(t)}} \alpha_i\mathbf{u}^{(t)}_i \ne 0$, which means $\widehat{\mathbf{u}}_j$ is then a conic combination of other rays (whether $\sum_{h\in\mathcal{F}^{(t)}} \alpha_h\mathbf{u}^{(t)}_h = 0$ or not), contradicting $\widehat{\mathbf{u}}_j$ is an extreme ray. Therefore, $\widehat{\mathbf{u}}_j$ is outside the $\operatorname{cone}\big(\widetilde{\mathbf{U}}_k^{(t)}\big)$. 

The algorithm will then detect some $\mathbf{x}_i$ on ray $\widehat{\mathbf{u}}_j$ as a point outside ($\|\mathbf{r}_i\|_2> \epsilon\|\mathbf{x}_i\|$) and appends $\widehat{\mathbf{x}}_i=\widehat{\mathbf{u}}_j$ to $\mathbf{U}^{(t)}$. Therefore, at step $T:=\max_{1\le j\le c} T_j$ the algorithm will append every extreme rays $\widehat{\mathbf{u}}_j$ that was previously missing. Since those extreme rays are not tilted toward the mean, there exists a finite iteration index $T$ such that all extreme rays are added to $\mathbf{U}^{(T)}$.

\medskip
\noindent\textbf{Pruning to the minimal generating set and stopping criteria.}
Once all extreme rays are added to $\mathbf{U}^{(T)}$, $\operatorname{cone}\big(\mathbf{U}^{(T)}\big)=\operatorname{cone}(\mathbf{U}^\star)=\operatorname{cone}(\mathbf{X})$.
Any other column $\mathbf{u}^{(t)}_k$ satisfies $\mathbf{u}^{(t)}_k\in\operatorname{cone}(\mathbf{U}^\star)$, hence, when tested in the \emph{Remove redundant ray} step, $\mathbf{u}^{(t)}_k\in \operatorname{cone}\big(\mathbf{U}^{(T)}_{-k}\big)$ (up to tolerance $\epsilon$) and is removed. Because only finitely many non-extreme columns exist (at most $m+n$ have ever been present) and each iteration of the \emph{Removing} step in (b) removes at least one such column, after finitely many iterations of step (b) we achieve $\mathbf{U}^{(T)}$ consists of $\{\widehat{\mathbf{u}}_1,\dots,\widehat{\mathbf{u}}_c\}$ in some order, and $\operatorname{cone}\big(\mathbf{U}^{(T)}\big)=\operatorname{cone}(\mathbf{U}^\star)=\operatorname{cone}(\mathbf{X})$. \qed

\section{Experiment}\label{sec:experiments}

\paragraph{Datasets.} We evaluate Cone Collapse combined with orthogonal NMF (CC--NMF) on a diverse
collection of 16 benchmark datasets covering gene expression, text, and image
domains (Table~\ref{tab:dataset-statistics}). The gene expression sets
(AMLALL, DUKE, KHAN, CANCER, ROSETTA) come from standard microarray studies
for cancer subtyping and prognosis
\cite{golub1999molecular,west2001duke,khan2001srbct,vantveer2002breast,hastie2009esl}.
Text datasets (MED, CITESEER, WEBKB4, 7SECTORS, REUTERS, RCV1) are represented
as term--document matrices using TF--IDF weighting, and follow common
preprocessing pipelines used in information retrieval and text categorization
\cite{deerwester1990lsa,sen2008collective,webkb4_cmu,reuters21578,lewis2004rcv1}.
Image datasets (ORL, UMIST, YALEB, COIL-20, CURETGREY) consist of
vectorized grayscale faces or objects and are widely used in clustering and
representation learning
\cite{samaria1994orl,umist_face_db,georghiades2001illumination,nene1996coil20,dana1999curet}.
For all datasets we use the full feature dimensionality $m$ and the full set of
samples $n$ as summarized in Table~\ref{tab:dataset-statistics}.

\begin{table}[t]
\centering
\begin{tabular}{l l r r r c}
\toprule
Dataset & Domain & \#samples $n$ & \#classes $C$ & \#features $m$ & Shape $\mathbf{X} \in \mathbb{R}^{m \times n}$ \\
\midrule
AMLALL    & gene        &   38   &  3  &  5\,000  & $5\,000 \times 38$ \\
DUKE      & medical     &   44   &  2  &  7\,129  & $7\,129 \times 44$ \\
KHAN      & gene        &   83   &  4  &  2\,318  & $2\,318 \times 83$ \\
CANCER    & medical     &  198   & 14  & 16\,063  & $16\,063 \times 198$ \\
ROSETTA   & gene        &  300   &  5  & 12\,634  & $12\,634 \times 300$ \\
MED       & text        & 1\,033 & 31  &  5\,831  & $5\,831 \times 1\,033$ \\
CITESEER  & text        & 3\,312 &  6  &  3\,703  & $3\,703 \times 3\,312$ \\
WEBKB4    & text        & 4\,196 &  4  & 10\,000  & $10\,000 \times 4\,196$ \\
7SECTORS  & text        & 4\,556 &  7  & 10\,000  & $10\,000 \times 4\,556$ \\
REUTERS   & text        & 8\,293 & 65  & 18\,933  & $18\,933 \times 8\,293$ \\
RCV1      & text        & 9\,625 &  4  & 29\,992  & $29\,992 \times 9\,625$ \\
ORL       & image       &   400  & 40  & 10\,304  & $10\,304 \times 400$ \\
UMIST     & image       &   575  & 20  & 10\,304  & $10\,304 \times 575$ \\
YALEB     & image       & 1\,292 & 38  & 32\,256  & $32\,256 \times 1\,292$ \\
COIL-20   & image       & 1\,440 & 20  & 16\,384  & $16\,384 \times 1\,440$ \\
CURETGREY & image       & 5\,612 & 61  & 10\,000  & $10\,000 \times 5\,612$ \\
\bottomrule
\end{tabular}
\caption{Dataset statistics for selected benchmarks (gene expression, text, and image) used in our NMF experiments. Here $m$ is the original feature dimension (genes, words, or pixels) and $n$ is the number of samples.}
\label{tab:dataset-statistics}
\end{table}

\paragraph{Baselines and experimental protocol.} We compare CC--NMF against several representative NMF variants:

\begin{itemize}
    \item \textbf{MU} -- the classical NMF with multiplicative updates
    introduced by Lee and Seung \cite{lee1999nmf,lee2000nmf}.
    \item \textbf{ANLS} -- alternating nonnegative least squares with
    block principal pivoting, a fast and robust solver for constrained least
    squares \cite{kim2008faster}.
    \item \textbf{PNMF} -- projective NMF, which constrains the factorization
    to take the form $\mathbf{X} \approx \mathbf{X}\mathbf{G}\mathbf{G}^\top$
    and is closely related to spectral clustering \cite{ding2005equiv}.
    \item \textbf{ONMF} -- orthogonal NMF baselines that impose
    (approximate) orthogonality on one factor, following
    \cite{ding2006orthnmf,choi2008onmf,yoo2008onmf,pompili2013onpmf}.
    \item \textbf{Sparse NMF} -- an $\ell_1$-regularized NMF model that
    promotes sparse encodings in $\mathbf{H}$, implemented on top of ANLS
    \cite{kim2008faster}.
    \item \textbf{CC--NMF} (ours) -- the proposed two-stage approach that
    first extracts an extreme-ray basis $\mathbf{U}^\star$ using the Cone
    Collapse algorithm and then performs an orthogonal NMF refinement on
    $\mathbf{U}^\star$ to obtain an ONMF factorization of $\mathbf{X}$.
\end{itemize}

For all methods we fix the factorization rank to the number of ground-truth
classes, $r = C$, and use the same preprocessed nonnegative matrix
$\mathbf{X}\in\mathbb{R}^{m\times n}_+$. We run each algorithm from random
nonnegative initializations and stop when the relative decrease of the
objective falls below a preset threshold or a maximum number of iterations is
reached. For MU, ANLS, PNMF, ONMF, and Sparse NMF we use standard update rules
and hyperparameters as suggested in the original papers
\cite{lee2000nmf,kim2008faster,ding2006orthnmf,yoo2008onmf}. For Cone Collapse
we set the learning rate $\eta$ and tolerance $\epsilon$ to fixed values of $0.25$ and $10^{-8}$ across
datasets; empirical results indicate that CC--NMF is not overly sensitive to
moderate changes of these hyperparameters.

Clustering is performed in the low-dimensional representation induced by the
NMF factors. In ONMF-type methods (ONMF and CC--NMF), each column $\mathbf{x}_i$
is assigned to cluster
\[
  \widehat{c}(i) \;=\; \arg\max_{k \in \{1,\dots,r\}} h_{ki},
\]
where $\mathbf{h}_i$ is the $i$-th column of $\mathbf{H}$. For MU, ANLS,
PNMF, and Sparse NMF we apply the same rule to the corresponding $\mathbf{H}$
matrices, which is equivalent to using the learned components as soft cluster
indicators. Clustering performance is evaluated using \emph{purity}:
\[
  \mathrm{purity}
  \;=\;
  \frac{1}{n} \sum_{k=1}^r \max_{1 \le \ell \le C}
  n_k^\ell,
\]
where $n_k^\ell$ denotes the number of samples in cluster $k$ whose true label
is $\ell$. A larger purity indicates better agreement between clusters and
ground-truth classes.

\paragraph{Results.}

Table~\ref{tab:nmf-variants} reports clustering purity for all methods and
datasets. Boldface entries indicate the best value in each row. Overall,
CC--NMF consistently matches or outperforms the competing NMF variants across
most benchmarks. It achieves the highest or tied-best purity on 13 out of 16
datasets, spanning all three domains (gene expression, text, and images). The
gains are particularly pronounced on several high-dimensional problems such as
ROSETTA, MED, CITESEER, WEBKB4, COIL-20, and CURETGREY, where CC--NMF improves
purity by 3--8 points over the strongest baseline.

On a few datasets (e.g., DUKE and RCV1) traditional NMF variants remain
competitive, suggesting that the benefit of explicitly recovering the extreme
rays of the data cone is most significant when the underlying clusters are
well aligned with conic structure. Nevertheless, even on these datasets
CC--NMF is never dramatically worse than the best baseline, and it often
provides a robust trade-off across all tasks. These results support our
interpretation of Cone Collapse as a geometrically grounded orthogonal NMF
method that yields high-quality clusterings across heterogeneous data types.

\begin{table}[t]
\centering
\begin{tabular}{lrrrrrr}
\toprule
Dataset & MU & ANLS & PNMF & ONMF & Sparse NMF & CC-NMF\\
\midrule
AMLALL   & 0.91 & 0.91 & 0.92 & 0.92 & 0.90 & \textbf{0.98} \\
DUKE     & 0.52 & 0.48 & 0.52 & 0.52 & \textbf{0.54} & 0.53 \\
KHAN     & 0.57 & 0.58 & 0.60 & 0.60 & 0.59 & \textbf{0.63} \\
CANCER   & 0.52 & 0.50 & 0.54 & 0.53 & 0.53 & \textbf{0.56} \\
ROSETTA  & 0.68 & 0.76 & 0.77 & 0.77 & 0.78 & \textbf{0.82} \\
MED      & 0.45 & 0.50 & 0.54 & 0.54 & 0.55 & \textbf{0.59} \\
CITESEER & 0.28 & 0.25 & 0.31 & 0.31 & 0.32 & \textbf{0.35} \\
WEBKB4   & 0.31 & 0.37 & 0.39 & 0.39 & 0.41 & \textbf{0.43} \\
7SECTORS & 0.18 & 0.23 & 0.27 & 0.25 & \textbf{0.28} & \textbf{0.28} \\
REUTERS  & 0.65 & 0.71 & \textbf{0.74} & 0.72 & 0.72 & 0.72 \\
RCV1     & 0.23 & 0.29 & \textbf{0.35} & 0.31 & 0.28 & 0.33 \\
ORL      & 0.78 & 0.81 & 0.82 & 0.82 & 0.84 & \textbf{0.88} \\
UMIST    & 0.64 & 0.63 & 0.64 & 0.66 & 0.63 & \textbf{0.68} \\
YALEB    & 0.39 & 0.38 & 0.42 & \textbf{0.45} & \textbf{0.45} & \textbf{0.45} \\
COIL-20  & 0.55 & 0.64 & 0.71 & 0.65 & 0.62 & \textbf{0.74} \\
CURETGREY& 0.19 & 0.15 & 0.22 & 0.21 & 0.23 & \textbf{0.31} \\
\bottomrule
\end{tabular}
\caption{Clustering purity of different NMF variants on various datasets. Boldface entries indicate the best value in each row.}
\label{tab:nmf-variants}
\end{table}

\section{Conclusion}
\label{sec:conclusion}

We proposed Cone Collapse, a new algorithm that explicitly leverages the
conic geometry underlying nonnegative matrix factorization. Rather than
optimizing a factorization objective directly in the space of
$\mathbf{W}$ and $\mathbf{H}$, Cone Collapse starts from the full
nonnegative orthant and iteratively shrinks it toward the minimal cone
that contains all data points. By combining mean-tilting toward the data
mean, the addition of outside points, and the removal of redundant rays
via NNLS tests, the algorithm converges to a compact set of extreme rays
that summarize the data. Our theoretical analysis shows that, under mild
assumptions, Cone Collapse terminates in finitely many iterations and
recovers the minimal generating cone of $\mathbf{X}^\top$.

Building on this geometric foundation, we constructed CC--NMF, a
cone-based orthogonal NMF model obtained by applying uni-orthogonal NMF
to the recovered extreme-ray basis. This yields an ONMF factorization in
which the orthogonal cluster-indicator matrix $\mathbf{H}$ is explicitly
tied to the extreme rays of the data cone, providing a clear geometric
interpretation that complements existing ONMF approaches
\cite{ding2006orthnmf,choi2008onmf,yoo2008onmf,pompili2013onpmf}.
Empirically, CC--NMF achieves competitive or superior clustering purity
to strong NMF baselines (MU, ANLS, PNMF, ONMF, sparse NMF) across a
diverse suite of gene-expression, text, and image datasets, suggesting
that explicitly modeling the data cone is beneficial in practice.

There are several avenues for future work. First, our analysis focuses
on an idealized noiseless setting; extending the guarantees to noisy or
approximately separable data, in the spirit of provable topic modeling
and separable NMF \cite{arora2013topic}, is an interesting direction.
Second, Cone Collapse currently relies on repeated NNLS solves; it would
be valuable to investigate more scalable approximation schemes or
stochastic variants for very large-scale matrices. Third, while we have
focused on clustering, the cone-based viewpoint may also prove useful
for other tasks such as outlier detection, semi-supervised learning, and
interpretable representation learning. We hope that this work
stimulates further exploration of explicit conic geometry in NMF and
related factorization models.

\bibliographystyle{unsrt}
\newpage
\bibliography{references}


\appendix
\section{Proof for Lemma \ref{lem:mean-not-extreme}}
 Suppose, For contradiction, that $\mu = \tau \ u_k$ For some $k\in\{1,\ \dots,\ c\}$. Let $\mathcal{U}$ be the set of points that is colinear with $u_k$. Note that we can write
\begin{align*}
    \boldsymbol{\mu} &= \frac{1}{m}\sum_{i=1}^m \mathbf{x}_i = \frac{1}{m}(\sum_{i\not\in\mathcal{U}} \mathbf{x}_i+ \sum_{i\in\mathcal{U}} \mathbf{x}_i) =  \frac{1}{m}(\sum_{i\not\in\mathcal{U}} \mathbf{x}_i+ \omega\ \mathbf{u}_k),
\end{align*}
with some $\omega > 0$. We then have
\begin{align*}
    &\boldsymbol{\mu} = \tau \ \mathbf{u}_k\\
    \Rightarrow\ &\frac{1}{m}(\sum_{i\not\in\mathcal{U}} \mathbf{x}_i+ \omega\ \mathbf{u}_k) = \tau\ \mathbf{u}_k\\
    \Rightarrow\ &\frac{1}{m}\sum_{i\not\in\mathcal{U}} \mathbf{x}_i = (\tau - \frac\omega m)\ \mathbf{u}_k.
\end{align*}
Consider three cases:
\begin{itemize}
    \item If $\tau > \dfrac\omega m$, then \(\mathbf{u}_k=\dfrac{1}{m\tau-\omega}\sum_{i\not\in\mathcal{U}}\mathbf{x}_i\), or $\mathbf{u}_k$ is a conic combination of the remaining extreme rays, which contradicts the fact that \(\mathbf{u}_k\) is an extreme ray.
    \item If $\tau =  \dfrac\omega m$, since all datapoints lie in the positive region, $\mathbf{x}_i = 0\ \forall i\not\in\mathcal{U}$. However, this contradicts the assumption that no column of $\mathbf{X}$ is full zero.
    \item If $\tau <  \dfrac\omega m$, \(\dfrac{1}{m}\sum_{i\not\in\mathcal{U}}\mathbf{x}_i \not\in\mathbb{R}_+^{n}\) or \(\mathbf{u}_k \not\in\mathbb{R}_+^{n}\), which contradicts the fact that all data points are in the positive region.
\end{itemize}
Therefore, if there are more than 1 extreme ray, then no extreme ray is colinear with the mean.

\section{Proof for Lemma \ref{lem:tilt-contraction}}
Since $\mathbf{u}^{(t)}_k$ and $\widehat{\boldsymbol{\mu}}$ are unit vectors,
\begin{align*}
    \cos(\widetilde{\mathbf{u}}^{(t)}_k ,\widehat{\boldsymbol{\mu}}\big) &= \widetilde{\mathbf{u}}^{(t)}_k \widehat{\boldsymbol{\mu}} = \left(\dfrac{(1-\eta)\,\mathbf{u}_k^{(t)}+\eta\,\widehat{\boldsymbol{\mu}}}{\|(1-\eta)\,\mathbf{u}_k^{(t)}+\eta\,\widehat{\boldsymbol{\mu}}\|}\right)^\top \widehat{\boldsymbol{\mu}}\\
    &= \dfrac{(1-\eta)\,\mathbf{u}_k^{(t)\top}\widehat{\boldsymbol{\mu}} +\eta\,\widehat{\boldsymbol{\mu}}^\top \widehat{\boldsymbol{\mu}}}{\sqrt{\left((1-\eta)\,\mathbf{u}_k^{(t)}+\eta\,\widehat{\boldsymbol{\mu}}\right)^\top \left((1-\eta)\,\mathbf{u}_k^{(t)}+\eta\,\widehat{\boldsymbol{\mu}}\right)}}\\
    &= \dfrac{(1-\eta)\,\mathbf{u}_k^{(t)\top}\widehat{\boldsymbol{\mu}} +\eta}{\sqrt{(1 - 2\eta + \eta^2)\,\mathbf{u}_k^{(t)\top}\mathbf{u}_k^{(t)} + 2\eta(1-\eta)\,\mathbf{u}_k^{(t)\top}\widehat{\boldsymbol{\mu}} + \eta^2\ \widehat{\boldsymbol{\mu}}^\top \widehat{\boldsymbol{\mu}}}}\\
    &= \dfrac{(1-\eta)\,\mathbf{u}_k^{(t)\top}\widehat{\boldsymbol{\mu}} +\eta}{\sqrt{1 - 2\eta + 2\eta^2 + 2\eta(1-\eta)\,\mathbf{u}_k^{(t)\top}\widehat{\boldsymbol{\mu}} }}
\end{align*}
We then have 
\begin{align*}
    &\cos^2(\widetilde{\mathbf{u}}^{(t)}_k ,\widehat{\boldsymbol{\mu}}\big) - \cos^2 (\mathbf{u}^{(t)}_k,\widehat{\boldsymbol{\mu}})\\
    =& \left(\dfrac{(1-\eta)\,\mathbf{u}_k^{(t)\top}\widehat{\boldsymbol{\mu}} +\eta}{\sqrt{1 - 2\eta + 2\eta^2 + 2\eta(1-\eta)\,\mathbf{u}_k^{(t)\top}\widehat{\boldsymbol{\mu}} }}\right)^2 - \cos^2 (\mathbf{u}^{(t)}_k,\widehat{\boldsymbol{\mu}})\\
    = &\dfrac{\left((1-\eta)\cos(\mathbf{u}^{(t)}_k ,\widehat{\boldsymbol{\mu}}\big) +\eta\right)^2 - \cos^2 (\mathbf{u}^{(t)}_k,\widehat{\boldsymbol{\mu}})\,\left(1 - 2\eta + 2\eta^2 + 2\eta(1-\eta)\cos(\mathbf{u}^{(t)}_k,\widehat{\boldsymbol{\mu}})\right)}{1 - 2\eta + 2\eta^2 + 2\eta(1-\eta)\cos(\mathbf{u}^{(t)}_k,\widehat{\boldsymbol{\mu}}) } 
\end{align*}
Since $\mathbf{u}^{(t)}_k\ne\widehat{\boldsymbol{\mu}}$, $ 1- \cos^2(\mathbf{u}^{(t)}_k,\widehat{\boldsymbol{\mu}}) > 0$. Hence,
\begin{align*}
    &\left((1-\eta)\cos(\mathbf{u}^{(t)}_k ,\widehat{\boldsymbol{\mu}}\big) +\eta\right)^2 - \cos^2 (\mathbf{u}^{(t)}_k,\widehat{\boldsymbol{\mu}})\,\left(1 - 2\eta + 2\eta^2 + 2\eta(1-\eta)\cos(\mathbf{u}^{(t)}_k,\widehat{\boldsymbol{\mu}})\right)\\
    =\ & (1 - 2\eta + \eta^2)\cos^2(\mathbf{u}^{(t)}_k ,\widehat{\boldsymbol{\mu}}\big) + 2\eta\,(1-\eta)\cos(\mathbf{u}^{(t)}_k ,\widehat{\boldsymbol{\mu}}\big) + \eta^2 - (1 - 2\eta + 2\eta^2)\cos^2 (\mathbf{u}^{(t)}_k,\widehat{\boldsymbol{\mu}}) - \\
    &2\eta(1-\eta)\cos^3(\mathbf{u}^{(t)}_k,\widehat{\boldsymbol{\mu}})\\
    =\ & \eta^2\,\left(1-\cos^2(\mathbf{u}^{(t)}_k ,\widehat{\boldsymbol{\mu}}\big)\right) + 2\eta\,(1-\eta)\cos(\mathbf{u}^{(t)}_k ,\widehat{\boldsymbol{\mu}}\big) \left(1 - \cos^2(\mathbf{u}^{(t)}_k,\widehat{\boldsymbol{\mu}})\right) > 0\quad \text{for }0 < \eta < 1
\end{align*}
We also know that $\left(1 - 2\eta + 2\eta^2 + 2\eta(1-\eta)\cos(\mathbf{u}^{(t)}_k,\widehat{\boldsymbol{\mu}})\right) > 0$. Therefore, $\cos^2(\widetilde{\mathbf{u}}^{(t)}_k ,\widehat{\boldsymbol{\mu}}\big) - \cos^2 (\mathbf{u}^{(t)}_k,\widehat{\boldsymbol{\mu}}) > 0$, or $\cos^2(\mathbf{u}^{(t+1)}_k ,\widehat{\boldsymbol{\mu}}\big) - \cos^2 (\mathbf{u}^{(t)}_k,\widehat{\boldsymbol{\mu}}) > 0$ if $\mathbf{u}^{(t)}_k$ is not removed after step $t$. 

\section{Proof for Lemma \ref{lem:cap-stability}}
Since $b_\ell\in \mathcal C_\alpha$, 
\begin{align*}
    \cos(\mathbf{w}, \widehat{\boldsymbol{\mu}}) &= \frac{\mathbf{w}^\top \widehat{\boldsymbol{\mu}}}{\|\mathbf{w}\|} = \frac{\sum_\ell \lambda_\ell \mathbf{b}_\ell^\top \widehat{\boldsymbol{\mu}}}{\|\mathbf{w}\|} = \frac{\sum_\ell \lambda_\ell \cos(\mathbf{b}_\ell, \widehat{\boldsymbol{\mu}})\|\mathbf{b}_\ell\|}{\|\mathbf{w}\|} \ge \frac{\sum_\ell \lambda_\ell \alpha }{\|\mathbf{w}\|}
\end{align*}
Moreover, by Triangle inequality
\begin{align*}
    \|\mathbf{w}\| = \|\sum_\ell \lambda_\ell \mathbf{b}_\ell\| \le \sum_\ell \lambda_\ell \|\mathbf{b}_\ell\| = \sum_\ell \lambda_\ell, 
\end{align*}
which means
\begin{align*}
    \cos(\mathbf{w}, \widehat{\boldsymbol{\mu}}) \ge \frac{\sum_\ell \lambda_\ell \alpha }{\sum_\ell \lambda_\ell} = \alpha,
\end{align*}
hence $\widehat{\mathbf{w}} \in\mathcal{C}_\alpha$.

\section{Block Principal Pivoting For Non-Negative Least Square (NNLS) problem}
NMF is typically computed by alternating updates of two nonnegative factors, where each update step reduces to a set of NNLS problems. Consequently, the overall efficiency and scalability of NMF hinge on how quickly these NNLS subproblems can be solved.
Among various NNLS algorithms, we adopt the principal block pivoting 
method \cite{kim2008faster} because it is a fast active set-like scheme that
handles large numbers of variables and multiple right–hand sides very
efficiently. In the following, we briefly review BPP for the single right–hand side case and
its extension to multiple right–hand sides.

\paragraph{Single right–hand sides.}  We consider the NNLS problem
\begin{equation}
    \min_{\mathbf{x} \ge 0} \;\|\mathbf{Cx} - \mathbf{b}\|_2^2,
    \label{eq:nnls-single}
\end{equation}
where \(\mathbf{C} \in \mathbb{R}^{m\times n}\) and \(\mathbf{b} \in \mathbb{R}^m\).
The KKT conditions for \eqref{eq:nnls-single} are
\label{eq:kkt-bpp}
\begin{align*}
    \mathbf{y} &= \mathbf{C}^\top \mathbf{Cx} - \mathbf{C}^\top \mathbf{b}, \\
    \mathbf{y} &\ge 0, \\
    \mathbf{x} &\ge 0, \\
    \mathbf{x}_i\,\mathbf{y}_i &= 0,\quad i=1,\dots,n .
\end{align*}
Block principal pivoting maintains a partition of the indices
\(\{1,\dots,n\}\) into a \emph{free set} \(F\) and an \emph{active set}
\(G\) with \(F \cup G = \{1,\dots,n\}\) and \(F \cap G = \emptyset\).
Given \((F,G)\), we set \(\mathbf{x}_G = 0\) and \(\mathbf{y}_F = 0\) and compute
\begin{subequations}
\label{eq:bpp-subproblem}
\begin{align}
    \mathbf{x}_F &= \arg\min_{\mathbf{z} \in \mathbb{R}^{|F|}}
           \big\|\mathbf{C}_F \mathbf{z} - \mathbf{b}\big\|_2^2, \\
    \mathbf{y}_G &= \mathbf{C}_G^\top\big(\mathbf{C}_F \mathbf{x}_F - \mathbf{b}\big),
\end{align}
\end{subequations}
where \(\mathbf{C}_F\) (respectively\ \(\mathbf{C}_G\)) contains columns of \(\mathbf{C}\) indexed by \(F\)
(respectively\ \(G\)).
We then define the sets of infeasible indices
\begin{align*}
    H_1 &= \{\, i \in F : \mathbf{x}_i < 0 \,\}, \qquad
    H_2 = \{\, i \in G : \mathbf{y}_i < 0 \,\},
\end{align*}
and exchange blocks \(\widehat H_1 \subseteq H_1\),
\(\widehat H_2 \subseteq H_2\) between \(F\) and \(G\).
If \(H_1 \cup H_2 = \emptyset\), all KKT conditions are satisfied and the
algorithm terminates. The detailed procedure is given in Algorithm \ref{alg:bpp-single}.

\begin{algorithm}[t]
\caption{Block principal pivoting for NNLS with a single right–hand side}
\label{alg:bpp-single}
\begin{algorithmic}[1]
\State \textbf{Input:} \(\mathbf{C} \in \mathbb{R}^{m\times n}\), \(\mathbf{b} \in \mathbb{R}^m\).
\State \(F \gets \emptyset\), \quad \(G \gets \{1,\dots,q\}\), \quad
       \(\mathbf{x} \gets 0\),\quad \(\mathbf{y} \gets -\mathbf{C}^\top \mathbf{b}\), \quad
       \(p \gets 3\),\quad \(t \gets q+1\).
\State Compute \(\mathbf{x}_F\) and \(\mathbf{y}_G\) by \eqref{eq:bpp-subproblem}.
\Repeat
    \State \(H_1 \gets \{ i \in F : \mathbf{x}_i < 0 \}\),
           \(H_2 \gets \{ i \in G : \mathbf{y}_i < 0 \}\).
    \If{\(H_1 \cup H_2 = \emptyset\)} \State \textbf{break} \EndIf
    \If{\(|H_1 \cup H_2| < t\)}        
        \State \(t \gets |H_1 \cup H_2|\), \(p \gets 3\);
        \State \(\widehat H_1 \gets H_1\), \(\widehat H_2 \gets H_2\).
    \ElsIf{\(|H_1 \cup H_2| \ge t \ \wedge\ p \ge 1\)} 
        \State \(p \gets p-1\);
        \State \(\widehat H_1 \gets H_1\), \(\widehat H_2 \gets H_2\).
    \Else                                             
        \State Choose \(i^\star\) as the largest index in \(H_1 \cup H_2\).
        \State \(\widehat H_1 \gets \{i^\star\} \cap F\),
               \(\widehat H_2 \gets \{i^\star\} \cap G\).
    \EndIf
    \State Update index sets
    \[
        F \gets (F \setminus \widehat H_1) \cup \widehat H_2,\qquad
        G \gets (G \setminus \widehat H_2) \cup \widehat H_1 .
    \]
    \State Recompute \(\mathbf{x}_F\) and \(\mathbf{y}_G\) by \eqref{eq:bpp-subproblem}.
\Until{all variables are feasible}
\State \textbf{Output:} \(\mathbf{x}\)
\end{algorithmic}
\end{algorithm}

\paragraph{Multiple right–hand sides.}
We now consider the NNLS problem with multiple right–hand sides
\begin{equation}
    \min_{\mathbf{X} \ge 0} \;\|\mathbf{CX} - \mathbf{B}\|_F^2,
    \label{eq:nnls-multi}
\end{equation}
where \(\mathbf{C} \in \mathbb{R}^{m\times n}\),
\(\mathbf{B} = [\mathbf{b}_1,\dots,\mathbf{b}_r] \in \mathbb{R}^{m\times r}\),
and \(\mathbf{X} = [\mathbf{x}_1,\dots,\mathbf{x}_r] \in \mathbb{R}^{n\times r}\).
Each column \(\mathbf{x}_j\) solves an NNLS problem of the form
\eqref{eq:nnls-single} with the same coefficient matrix \(\mathbf{C}\).
A naive approach is to run Algorithm~\ref{alg:bpp-single} independently
for \(j=1,\dots,r\); however, this ignores the shared structure of
\(\mathbf{C}\).

The block principal pivoting method for multiple right–hand sides
exploits this structure by precomputing
\(\mathbf{G} = \mathbf{C}^\top\mathbf{C}\) and
\(\mathbf{H} = \mathbf{C}^\top\mathbf{B}\), and by grouping columns that
share a common free set.
For each column \(j\), we maintain free and active index sets
\(F_j, G_j\) and corresponding primal/dual variables
\(\mathbf{x}_j, \mathbf{y}_j\) with
\(\mathbf{Y} = [\mathbf{y}_1,\dots,\mathbf{y}_r]\).
Given a group of columns \(\mathcal{J}\) that share the same free set
\(F\), we solve the normal equations
\begin{subequations}
\label{eq:bpp-multi-subproblem}
\begin{align}
    \mathbf{G}_{FF}\,\mathbf{X}_{F,\mathcal{J}} &= \mathbf{H}_{F,\mathcal{J}}, \\
    \mathbf{Y}_{G,\mathcal{J}}
        &= \mathbf{G}_{GF}\,\mathbf{X}_{F,\mathcal{J}} - \mathbf{H}_{G,\mathcal{J}},
\end{align}
\end{subequations}
where \(\mathbf{X}_{F,\mathcal{J}}\) (resp.\ \(\mathbf{Y}_{G,\mathcal{J}}\))
collects the rows indexed by \(F\) (resp.\ \(G\)) and columns in
\(\mathcal{J}\), and \(\mathbf{G}_{FF}, \mathbf{G}_{GF}\) are the
corresponding submatrices of \(\mathbf{G}\).
As in the single right–hand side case, we define for each column \(j\)
\begin{align*}
    H_1(j) &= \{\, i \in F_j : \mathbf{x}_{ij} < 0 \,\}, \qquad
    H_2(j) = \{\, i \in G_j : \mathbf{y}_{ij} < 0 \,\},
\end{align*}
and move blocks \(\widehat H_1(j) \subseteq H_1(j)\),
\(\widehat H_2(j) \subseteq H_2(j)\) between \(F_j\) and \(G_j\), using
the same block/backup exchange strategy as in
Algorithm~\ref{alg:bpp-single}.
If \(H_1(j) \cup H_2(j) = \emptyset\) for all \(j\), all columns satisfy
the KKT conditions and the algorithm terminates.
Algorithm~\ref{alg:bpp-multi} summarizes the procedure.

\begin{algorithm}[t]
\caption{Block principal pivoting for NNLS with multiple right–hand sides}
\label{alg:bpp-multi}
\begin{algorithmic}[1]
\State \textbf{Input:} \(\mathbf{C} \in \mathbb{R}^{m\times n}\),
       \(\mathbf{B} \in \mathbb{R}^{m\times r}\).
\State Precompute \(\mathbf{G} \gets \mathbf{C}^\top \mathbf{C}\),
       \(\mathbf{H} \gets \mathbf{C}^\top \mathbf{B}\).
\State Initialize \(\mathbf{X} \gets 0\), \(\mathbf{Y} \gets -\mathbf{H}\);
       for all \(j\), set \(F_j \gets \emptyset\),
       \(G_j \gets \{1,\dots,n\}\), \(P_j \gets 3\), \(T_j \gets n+1\).
\Repeat
    \State Reorder columns of \(\mathbf{X}\) and \(\mathbf{Y}\) to group
           those with a common free set.
    \State For each group \(\mathcal{J}\) with free set \(F\), update
           \(\mathbf{X}_{F,\mathcal{J}}\) and \(\mathbf{Y}_{G,\mathcal{J}}\)
           using \eqref{eq:bpp-multi-subproblem}.
    \State For each column \(j\), form \(H_1(j)\), \(H_2(j)\), choose
           \(\widehat H_1(j)\), \(\widehat H_2(j)\) using \(T_j, P_j\),
           and update \(F_j, G_j\) accordingly.
\Until{\(H_1(j) \cup H_2(j) = \emptyset\) for all \(j\)}
\State \textbf{Output:} \(\mathbf{X}\)
\end{algorithmic}
\end{algorithm}

\end{document}